\pdfoutput=1

\documentclass[11pt]{article}

\usepackage[]{acl}
\usepackage{float}

\usepackage{times}
\usepackage{latexsym}
\usepackage{booktabs}
\usepackage{amsthm}
\usepackage{amsmath}
\usepackage{amsfonts}
\usepackage{multirow}

\usepackage{todonotes}
\usepackage{xcolor}
\usepackage{enumitem}
\usepackage{listings}
\usepackage{amssymb}
\usepackage{pifont}
\usepackage{ifthen}
\newcommand{\cmark}{\ding{51}}
\newcommand{\xmark}{\color{codepurple}\ding{55}}

\usepackage[T1]{fontenc}

\usepackage[utf8]{inputenc}

\usepackage{microtype}

\definecolor{title}{HTML}{687C87} 

\newboolean{is_include_image_for_fusion_placement}
\setboolean{is_include_image_for_fusion_placement}{false}

\DeclareRobustCommand{\VAN}[3]{#2} 

\title{\vspace{-18px}{\LARGE Arctic-TILT}\vspace{3px}\\Business Document Understanding at Sub-Billion Scale\vspace{13px}}

\author{
{\normalfont Łukasz Borchmann}\thanks{\, \, See Appendix~\ref{sec:contributions} for contributions.} \\\And
{\normalfont Michał Pietruszka$^{\dag}$} \\\And 
{\normalfont Wojciech Jaśkowski} \\\And
{\normalfont Dawid Jurkiewicz} \\\AND
{\normalfont Piotr Halama} \\\And
{\normalfont Paweł Józiak$^{\ddag}$} \\\And 
{\normalfont Łukasz Garncarek} \\\And
{\normalfont Paweł Liskowski} \\\AND
{\normalfont Karolina Szyndler} \\\And 
{\normalfont Andrzej Gretkowski} \\\And 
{\normalfont Julita Ołtusek} \\ \And
{\normalfont Gabriela Nowakowska$^{\S}$} \\\AND
{\normalfont Artur Zawłocki} \\ \And 
{\normalfont Łukasz Duhr} \\\And
{\normalfont Paweł Dyda} \\\And 
{\normalfont Michał Turski$^{\S}$} \vspace{14px}\\ \AND 
Snowflake \\
  \texttt{name.surname@snowflake.com} \\\vspace{2px}\AND
  \(^{\S}\){\normalfont \normalsize Adam Mickiewicz University}\And
  $^{\dag}${\normalfont\normalsize Jagiellonian University}\And $^{\ddag}${\normalfont\normalsize Warsaw University of Technology}
}

\definecolor{codegreen}{HTML}{29B5E8}
\definecolor{codegray}{HTML}{7254A3}
\definecolor{codepurple}{HTML}{D45B90}
\definecolor{backcolour}{HTML}{F7F7F7}

\lstdefinestyle{mystyle}{
    language=Python,
    backgroundcolor=\color{backcolour},   
    commentstyle=\color{codegreen},
    keywordstyle=\color{codepurple},
    numberstyle=\tiny\color{gray},
    stringstyle=\color{codepurple},
    basicstyle=\ttfamily\small,
    breakatwhitespace=false,         
    breaklines=true,                 
    captionpos=b,                    
    keepspaces=true,                 
    numbers=left,                    
    numbersep=25pt,                  
    showspaces=false,                
    showstringspaces=false,
    showtabs=false,                  
    tabsize=2,
    framexleftmargin=16pt,
    framexrightmargin=16pt,
    framextopmargin=6pt,
    framexbottommargin=6pt, 
    frame=tb, framerule=0pt,
}

\begin{document}
\maketitle

\begin{abstract}
The vast portion of workloads employing LLMs involves answering questions grounded on PDF or scan content.
We introduce the Arctic-TILT 
achieving
accuracy on par with models 1000$\times$ its size on these use cases. It can be fine-tuned and deployed on a single 24GB GPU, lowering operational costs while processing Visually Rich Documents with up to 400k tokens. 
The model establishes state-of-the-art results on seven diverse Document Understanding benchmarks, 
as well as provides reliable confidence scores and quick inference, which are essential for processing files 
in large-scale or time-sensitive enterprise environments.

\end{abstract}

\begin{figure}[bh]
    \centering
    \includegraphics[width=0.78\linewidth]{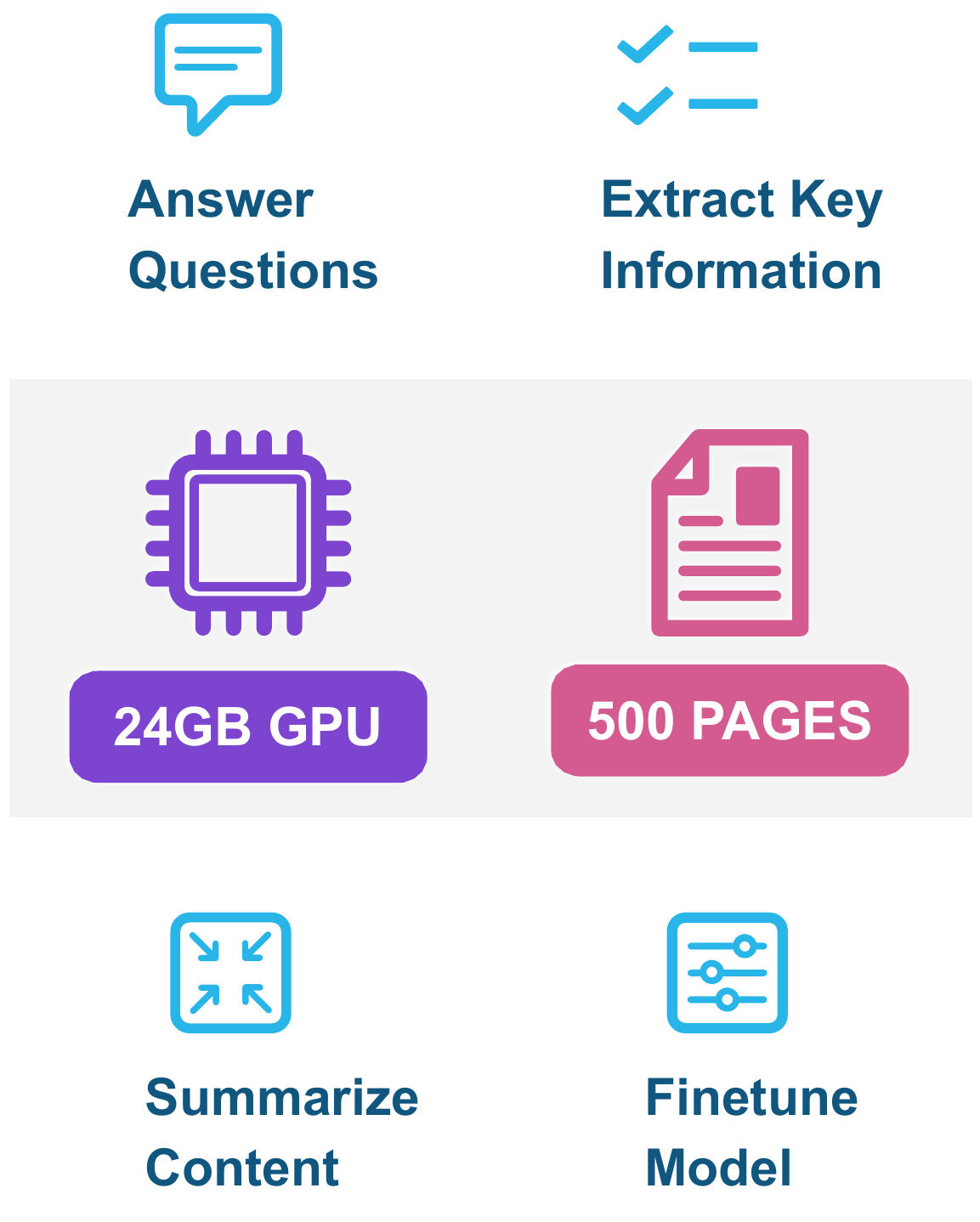}
    \caption{Arctic-TILT consumes long, richly formatted PDFs given a single, cost-efficient GPU and can produce their summary, answer questions, and extract values, outperforming vastly heavier LLMs and LVLMs.\vspace{-16px}}
    \label{fig:hero}
\end{figure}

\section{Introduction}  

General-purpose LLMs and their multi-modal counterparts provide a crucial advantage in process automation: they can be applied immediately, eliminating the expensive and time-consuming efforts of creating dedicated system architecture and model development.
Though they are suitable choices for prototyping and building proof-of-concept solutions, once the case is validated, it becomes essential to consider the demands of real-world deployments, such as cost-efficiency \cite{fu2024tinytitanssmallerlarge,ong2024routellmlearningroutellms}, fine-tunability \cite{liu2022fewshotparameterefficientfinetuningbetter}, and ensuring accurate confidence calibration \Citep{VanLandeghem2024phdthesis}.

We consider these issues in the context of Document Understanding (DU), where it is commonly required to integrate textual, layout and graphical clues to obtain the required information
and introduce the Arctic-TILT, designed to 
address the needs of broad-use deployments, cost efficiency, and domain adaptations for a
fraction of the cost of the leading models.
The proposed solution achieves state-of-the-art accuracy on business and long document benchmarks of MP-DocVQA \cite{tito2023hierarchicalmultimodaltransformersmultipage}, DUDE \Citep{vanlandeghem2023documentunderstandingdatasetevaluation}, Kleister NDA and Charity \cite{DBLP:journals/corr/abs-2105-05796}, ArXiv-Lay and PubMed-Lay \cite{nguyen2023loralaymultilingualmultimodaldataset}, and remains competitive with orders of magnitude larger models on other document VQA datasets.

\begin{table}[t]
    \small
    \centering
    \begin{tabular}{p{0.39\linewidth}p{0.48\linewidth}}
    \toprule
    TILT & Arctic-TILT  \\
    \midrule
    \multicolumn{2}{l}{\textit{Vision Encoding and its Fusion with Text}}\vspace{1mm} \\
    sum of text \& image & fusion by tensor product  \\
    first layer only & every encoder layer \\
    \midrule
    \multicolumn{2}{l}{\textit{Pretraining and Finetuning}}\vspace{1mm} \\
    400k steps of adaptation & 900k steps of adaptation \\
    SFT on 4 datasets & SFT on 17 datasets \\
    \midrule
    \textit{Transformer}\vspace{1mm} \\
    dense attention, vanilla & sparse attention, SLED \\
    max 9k tokens & max 400k tokens \\
    basic optimization & heavy optimization \\
    \bottomrule
    \end{tabular}
    \caption{Comparison of TILT and Arctic-TILT.}
    \label{tab:comparison}
\end{table}

\section{Related Works}

The appearance of LLMs is a result of trends that have also been strongly visible in the DU field in recent years.

Traditionally, tasks like table or information extraction from Visually Rich Documents were broken down into separate steps, such as form recognition, field detection, and value extraction \cite{MTL_FoUn,medvet2011probabilistic,6628784,peanho2012semantic,tian2016detectingtextnaturalimage,le2019deeplearningapproachreceipt,baek2019characterregionawarenesstext,holt-chisholm-2018-extracting,Wang2021TowardsRV,Carbonell2019TreyNetAN}.
Each of these steps typically required distinct models or heuristics and processing pipelines that were later approached in a more end-to-end manner employing graph-based approaches \citep[\textit{inter alia}]{liu2019graphconvolutionmultimodalinformation,hwang2021spatialdependencyparsingsemistructured,9412927,wang2024docgraphlmdocumentalgraphlanguage}.
%

Ultimately, the DU field has converged on formulating tasks in a unified text-to-text format due to its robustness in handling various problems, which LLMs align well with due to their generic input-output format  \cite{mathew2021docvqadatasetvqadocument, mathew2021infographicvqa, borchmann2021due}. 
Although this approach appears elegant and its ease of application makes it appealing for industrial-scale implementation, treating documents as pure text is often insufficient, particularly where layout-intensive aspects dominate. 
Hence, there has been a recent surge in extending LLMs with visual \cite{li2023blip2bootstrappinglanguageimagepretraining, wu2023visualchatgpttalkingdrawing}, layout modality \cite{Fujitake2024LayoutLLMLL}, or both \cite{Mao2024VisuallyGG, li2024enhancingvisualdocumentunderstanding, tang2023unifyingvisiontextlayout} to better capture the nuances of document structures and improve performance on layout-intensive tasks.

A separate line of work approaches DU problems using vision-only models \cite{DBLP:journals/corr/abs-2111-15664,kim2022donut,lee2023pix2structscreenshotparsingpretraining,beyer2024paligemmaversatile3bvlm}, assuming one can address the problem without specific architectural biases. However, models with textual input outperform them with a notable example of GPT-4 Vision that benefits from the availability of OCR-recognized text \cite{borchmann2024notesapplicabilitygpt4document}.

Despite these advancements and the significant benefits the scale of LLMs offers, we provide arguments for smaller, problem-specific models, similarly to \citet{fu2024tinytitanssmallerlarge,zhao2024loraland310finetuned} and focus on cost-efficient deployment \cite{ong2024routellmlearningroutellms}. 

\begin{figure}[t]
    \centering
    \includegraphics[width=0.97\linewidth,trim={0.7cm 1.2cm 0.7cm 1.2cm},clip]{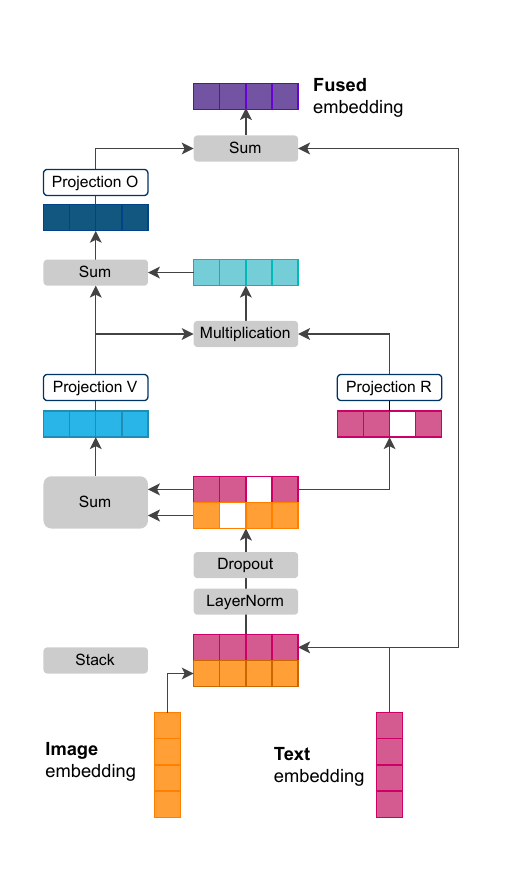}
    \caption{Arctic-TILT modality fusion. It can be seen as attention with role vector \cite{tensortproduct} simplified concerning we calculate it over a pair of aligned text and image tokens.}
    \label{fig:fusion}
\end{figure}

\section{Arctic-TILT}

Our starting point is the TILT encoder-decoder model, built upon T5 \cite{2020t5} by (1) extending its sequential positional bias with an attention bias based on relative horizontal and vertical distances between each pair of tokens, and (2) adding contextualized image embeddings that cover the token image region semantics in the context of its entire visual neighborhood \cite{tilt}.

We further enhance TILT's performance and remove limitations. Specifically, we propose novel modality fusion, introduce attention sparsity, enhance training recipe, and optimize training and inference (Table~\ref{tab:comparison}). Improved variant of the model is referred to as the Arctic-TILT.


\subsection{Fusion of Text and Vision}  

The TILT model's unique approach to combining visual and textual semantics involves summing word embeddings and RoI-pooled representations of the word's bounding box with a variant of the U-Net network used as an image encoder. Visual and textual features are integrated once, immediately after embedding both inputs.

We begin by replacing the post-embedding summation of text and vision modalities from the original TILT with our proposed mechanism, integrated within each transformer block.

\paragraph{Fusion by Tensor Product.} In contrast to \citet{tilt}, we opt for the fusion of modalities inspired by tensor product representations \cite{SMOLENSKY1990159,10.1007/978-1-4471-2063-6_110} and their Hadamard approximation \cite{tensortproduct}. Specifically, given the text and image embeddings $t, i \in \mathbb{R}^{d}$, we calculate the fused embedding with:
$$\operatorname{Fuse}(t, i) = O(V(t+i) \odot (\mathbf{1} + Rt)) + t$$
\noindent where $V$, $R$, and $O$ are $\mathbb{R}^{d\times d}$ trainable parameters. In practice, we use a variant of this mechanism with additional layer norm and dropout, as depicted in Figure~\ref{fig:fusion} and Listing~\ref{lst:fusion}.


\begin{figure}[t]
    \centering\hspace{-20px}
    \includegraphics[width=0.95\linewidth]{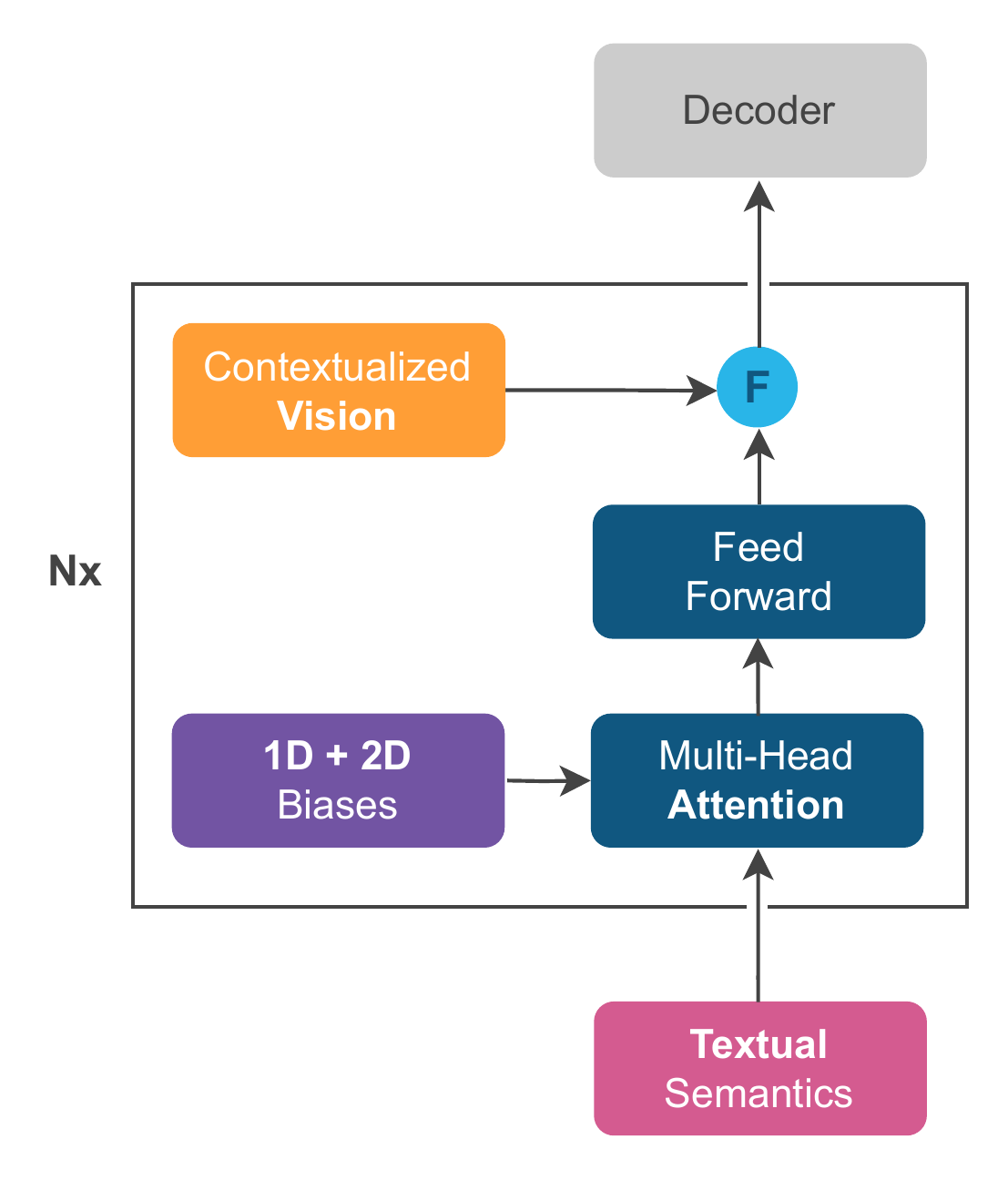}
    \caption{
    The Arctic-TILT encoder block combines \textit{Contextualized Vision} from U-Net and \textit{Textual Semantics} from input embeddings through \textit{Fusion} (F) operation. 
    The \textit{Multi-Head Attention} is augmented with \textit{1D and 2D positional biases} to capture spatial and sequential arrangement. This procedure is repeated in each layer (\textit{Nx}), allowing 
    to process integrated information further.
    }
    \label{fig:architecture}
\end{figure}

\paragraph{Module placement.}
\ifthenelse{\boolean{is_include_image_for_fusion_placement}}{
    \begin{figure}[h]
        \centering
        \begin{tabular}{c}
           \includegraphics[width=0.8\linewidth,trim={0.0cm 0.9cm 0.0cm 0.1cm},clip]{assets/fusion-applica-mix.png}\\
           \includegraphics[width=0.8\linewidth,trim={0.0cm 0.85cm 0.0cm 0.5cm},clip]{assets/fusion-buc.png} \\
           \includegraphics[width=0.8\linewidth,trim={0.0cm 1.45cm 0.0cm 0.5cm},clip]{assets/fusion-docvqa.png} \\
           \includegraphics[width=0.8\linewidth,trim={0.0cm 0.0cm 0.0cm 0.6cm},clip]{assets/fusion-infographicsvqa.png}

        \end{tabular}
        
        \caption{Downstream finetuning with different fusion setups. Two internal benchmarks (first and second), followed by DocVQA (third) and InfographicsVQA (fourth). Y-axis is ANLS.}
        \label{fig:fusion-placement}
    \end{figure}
}{
}

Having described the mechanism of integrating two modalities together, the question arises on the positioning of the fusion module within the 
transformer block. 

We found that placing the fusion module after FFNs is most beneficial as the fusion results are directly fed to the next layer (Figure~\ref{fig:architecture}). 
Additionally, by applying fusion after every encoder layer, we mitigate the vanishing gradient effect and enable the model to focus on different visual features as its comprehension of the document improves.

\subsection{Long Context Support}
Concerning the product-oriented nature of our work, it is essential to cover a significant fraction of real-world documents of potentially arbitrary lengths while operating within limited resources.

The outlined optimizations are guided by the need to handle as much context as possible on widely available A10 and L4 GPUs equipped with 24GB vRAM.
We assume a single-GPU setup and measure the impact of applied techniques and architectural changes on the maximum context length used during the finetuning and inference under this memory budget.

\paragraph{Chunked processing.} 
To address the quadratic complexity of self-attention computation in the encoder, we employ a variant of fusion-in-decoder~\cite{dejong2023fidofusionindecoderoptimizedstronger}, also known as blockwise encoding~\cite{pietruszka2022sparsifyingtransformermodelstrainable} or SLED~\cite{ivgi2022efficient} with zero chunk padding. This method restricts the encoder attention matrix to a neighborhood around its diagonal with a bounded width. Without padding, this results in a block diagonal matrix, reducing the number of non-zero attention weights to a linear quantity relative to the input sequence length.

Specifically, for a fixed \emph{core chunk length} \(c\), and overlap size \(o\), being hyperparameters of our model, prefix of length \(l\) and total input size \(C = n\cdot c\), we build a sequence of chunks in the following manner: chunk number \(1\) is  filled with \(l\) tokens of prefix, and with tokens \(0, .., c-l\) of the input. Say chunk number \(i\) already used input tokens of up to \(t\). Chunk number \(i+1\) starts with \(l\) tokens of prefix, followed with tokens \(t-o+1, t-o+2, ..., t-o+c-l\) of the input. 
In practice, we found a core chunk length of 1024 and no overlap to perform robustly across most tasks (see Appendix~\ref{sec:hparams}).

\begin{figure}[t]
    \centering
    \hspace{-0.6cm}\includegraphics[width=1\linewidth,trim=1cm 1.1cm 0.7cm 1.1cm,clip]{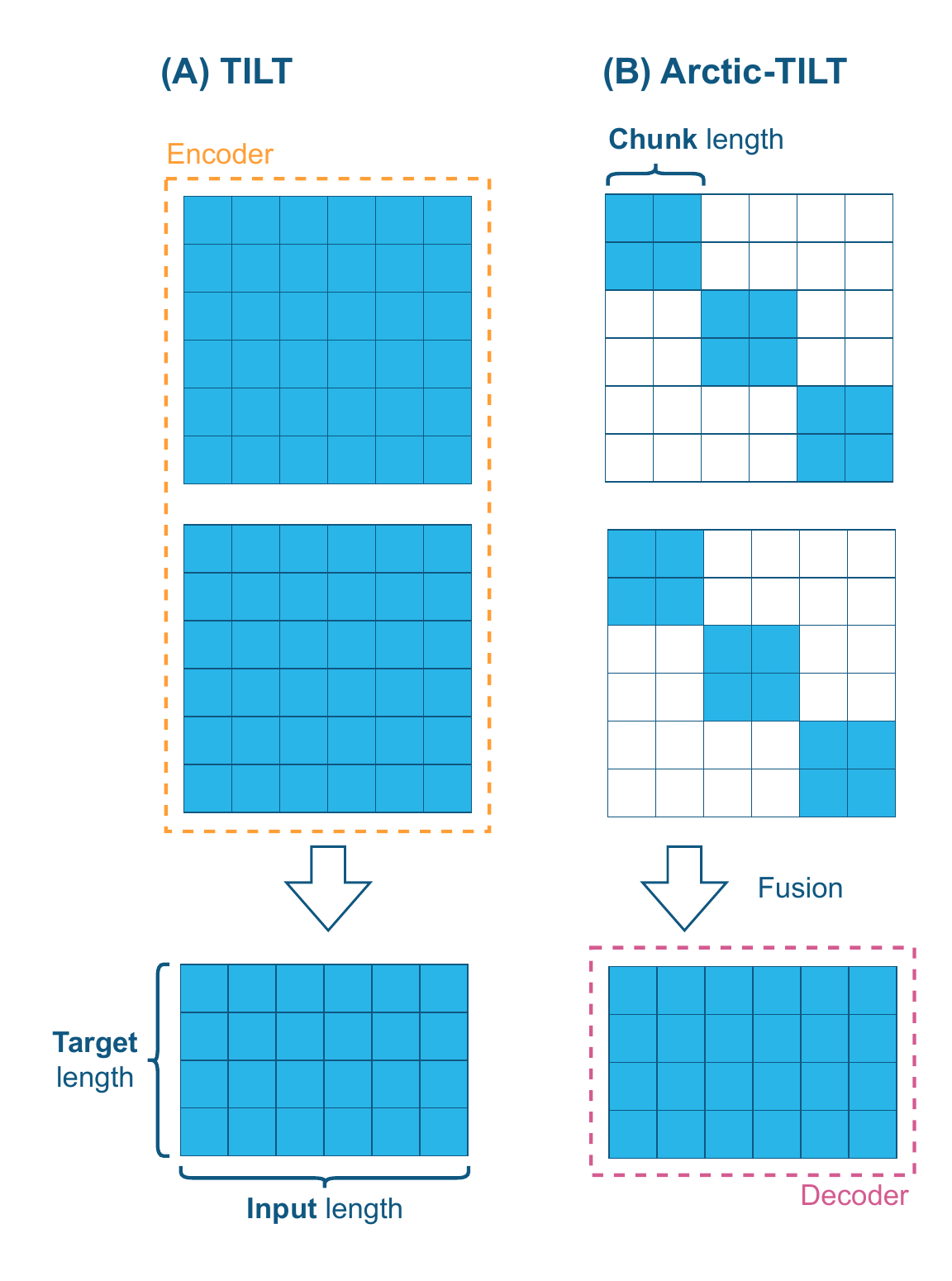}
    \caption{An illustration of sparse attention matrices assuming a two-layer encoder and decoder. 
    The original TILT (A) consumes the complete input at once, in contrast to Arctic-TILT (B) with blockwise attention}
    \label{fig:sparsity}
\end{figure}

The resulting sequences are passed through the encoder separately. The encoder outputs are then recombined by removing the prefix tokens embeddings from all but the first chunk and concatenating the results. The resulting encoder output is passed to the decoder (see Figure~\ref{fig:sparsity}).

Since the underlying attention implementation is already memory efficient, this technique improves the computational efficiency in training and inference, resulting in a 10-fold increase in input length during training and inference.

\paragraph{Nested stack checkpointing.} Implementing gradient checkpointing over the entire 24-layer encoder stack reduces the memory required for activations. Only the last layer's activations are stored, which are necessary for the decoder. Consequently, memory savings are significant here, as the requirement for processing 1M tokens reduced from 96GB to merely 4GB for the encoder part, albeit at the cost of an additional forward pass through the encoder. This improvement allowed to quadruple the input length for training. 

\paragraph{Random chunks.} While our modifications effectively handle the encoder's memory limitations, the length of concatenated chunk embeddings can still cause the decoder cross-attention to exceed available memory. 
Technically, the model can handle 230k tokens during the training, which was further addressed with a simple method that allows for extending the length of the document while also positively impacting the scores. We randomly discard chunks, effectively exposing the model to the different parts of longer documents across epochs. The first chunk, typically containing the initial pages, is always preserved to provide essential context.



\vspace{1em}
In addition to primary techniques, we employ several other optimizations. Specifically, we use mixed precision training with \textit{bfloat16} while turning off weight caching to save RAM, leading to a 2$\times$ improvement in inference input length. Secondly, by recomputing projections for each decoder layer instead of using the key-value cache, we extend the maximum inference context to 389k tokens. Next, we optimize training by offloading the decoder's activations needed for backpropagation from GPU to CPU, minimizing peak memory usage of the GPU by increasing processing time. 
Finally, implementing memory-efficient attention reduces the memory overhead of the attention mechanism \cite{rabe2022selfattentiondoesneedon2}.



\vspace{1em}\noindent 
Ultimately, our optimizations culminate in significant memory usage improvements, allowing us to effectively train and deploy Arctic-TILT for documents up to 500 pages\footnote{Specifically, 390k input tokens with an output of 128 tokens,  corresponding to 780 tokens per page on average.} on a single 24GB GPU. The step-by-step summary is studied in Table~\ref{tab:memory}.

\begin{table}[]
    \small
    \centering
    \begin{tabular}{lrr}
    \toprule
    & Inference & Training \\
    \midrule
    Vanilla TILT & 9k & 4k \\
    \quad + attention sparsity & 87k & 41k  \\
    \quad + mixed precision & 179k & 51k \\
    \quad + memory efficient attention & 183k & 56k \\
    \midrule
    \textit{Inference-only optimizations} \\
    \quad + no cross-attention KV cache & 389k & \\
    \midrule
    \textit{Training-only optimizations} \\
    \quad + nested checkpointing & & 230k \\
    \quad + CPU offloading & & 256k \\
    \quad + random chunks & & 389k \\
    \bottomrule
    \end{tabular}
    \caption{Max input length (tokens) consumed during training and inference given single 24GB GPU. Tested for documents up to 500 pages (389k tokens). 
    }
    \label{tab:memory}
\end{table}


\begin{table*}[ht]
    \small
    \centering
    \bgroup
    \setlength\tabcolsep{1.5em}
    \def\arraystretch{1.2}%
    \begin{tabular}{lcc|llc}
    \toprule
    Dataset & Industrial & Multipage & \multicolumn{2}{c}{State-of-the-Art} & Arctic-TILT \\
    \midrule
    MP-DocVQA & \cmark & \cmark & GRAM & 80.3 & \textbf{81.2} \\
    Kleister Charity & \cmark & \cmark & LAMBERT & 83.6 & \textbf{88.1} \\
    Kleister NDA & \cmark & \cmark & ERNIE-Layout & 88.1 & \textbf{94.3} \\
    DUDE & \cmark / \xmark & \cmark & GPT-4Vt + OCR & 53.9 & \textbf{58.1} \\
    MMLongBench-Doc$^{\dag}$ & \cmark / \xmark & \cmark & GPT-4o & \textbf{42.8} & 25.8 \\
    SlideVQA &  \xmark & \cmark & GPT-4Vt + OCR & \textbf{57.3} & 55.1 \\
    ArXiv-Lay & \xmark & \cmark & BigBird-Pegasus+Layout & 41.2  & \textbf{44.4} \\
    PubMed-Lay & \xmark & \cmark & BigBird-Pegasus+Layout & 42.1 & \textbf{44.8} \\
    DocVQA & \cmark & \xmark & InternVL 2.0 Pro & \textbf{95.1} & 90.2 \\
    VQA-CD & \cmark & \xmark & QALayout & 42.5 & \textbf{90.7} \\
    InfographicsVQA & \xmark & \xmark & InternVL 2.0 Pro & \textbf{86.8} & 57.0 \\
    \bottomrule
    \end{tabular}
    \egroup
    \caption{Arctic-TILT compared to the previous state-of-the-art. Our model remains competitive despite having less than 1B parameters and excels when input is a long, business document. We use the original metrics for each dataset, i.e., F1 for Kleisters, Accuracy for MMLongBench-Doc, EM for SlideVQA, ROUGE-L for ArXiv-Lay and PubMed-Lay, and ANLS for the remaining tasks; $^{\dag}$ denotes zero-shot evaluation.}
    \label{tab:results}
\end{table*}



\subsection{Pretraining and Finetuning}
The training process began with a self-supervised pretraining phase using the pretrained T5 large model~\cite{2020t5}. Following the introduction of TILT architecture changes, which included U-Net \cite{unet2015} and 2D biases, as well as text-vision post-fusion, the model underwent further self-supervised pretraining for a total of 900k steps based on documents from the CCpdf \cite{ccpdf} and OCR-IDL \cite{ocridl}. These two large-scale, publicly available PDF resources come with OCR results from Tesseract and Amazon Textract, respectively. 

Finally, the model was fine-tuned on QA and KIE datasets. In this phase, we increase the number of supervised datasets to 17, compared to TILT's original choice of four. The datasets chosen represent critical aspects of DU tasks, including, but not limited to, forms, financial reports, charts, invoices, insurance documents, contracts, and legal documents (detailed in Appendix~\ref{appendix:dataset}). 

\section{Experiments}

We start by examining diverse benchmarks in the Document Understanding field, focusing on these closer to enterprise applications due to the domain of the included documents and the presence of multi-page inputs that systems encounter in real-world applications (see Table~\ref{tab:results}).

Hyperparameters used during finetuning were subject to optimization outlined in Appendix~\ref{sec:hparams}.

\subsection{Document Visual QA and KIE}\label{sec:qa_kie}


Regarding the model's intended use for processing unstructured documents, Document Visual Question Answering and Key Information Extraction tasks appear best suited for assessing  performance. 

\paragraph{Multi-page.} Arctic-TILT excels when input is a long, business document. In the case of DUDE \Citep{vanlandeghem2023documentunderstandingdatasetevaluation} consisting of PDFs of up to 25 pages, sourced across industries, we outperform GPT-4 Vision Turbo by 4 points and Gemini 1.5 Pro by 12 points.  Qualitatively, Arctic-TILT not only outperforms GPT-4Vt in handling non-answerable questions but also exceeds state-of-the-art models in list and abstractive analysis tasks, an example showing complex data handling skills.

\begin{figure}[t]
    \centering
    \includegraphics[width=\linewidth]{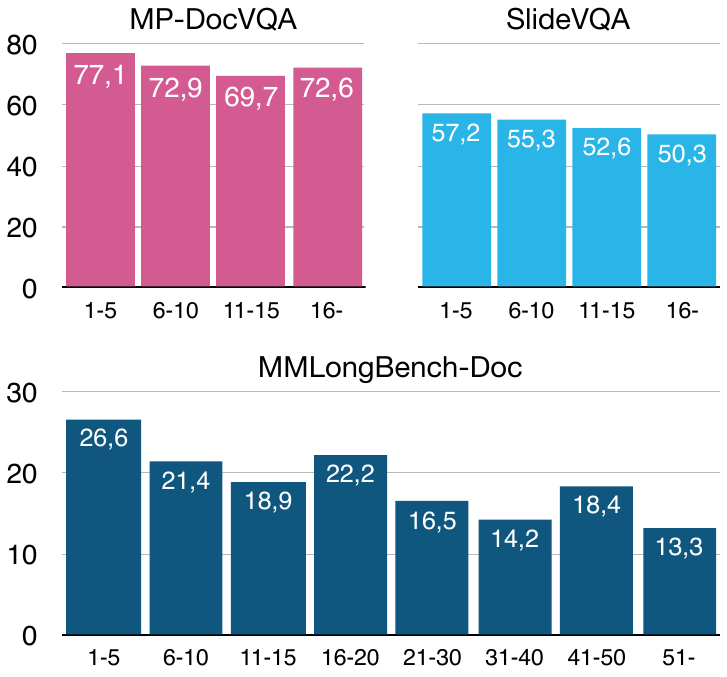}
    \caption{Scores of Arctic-TILT on MP-DocVQA, SlideVQA, and MMLongBench-Doc, depending on the evidence location (buckets of five pages).} 
    \label{fig:context}
\end{figure}

Similarly, we establish state-of-the-art performance on MP-DocVQA \cite{tito2023hierarchicalmultimodaltransformersmultipage} dataset consisting of questions posed to industrial documents of up to 20 pages, outperforming GRAM \cite{blau2024gramglobalreasoningmultipage} by 1 point. 
Further, we were able to surpass the long-standing score of LAMBERT \cite{Garncarek_2021} on Kleister Charity \cite{DBLP:journals/corr/abs-2105-05796}, with documents reaching above 300 pages, by 4 points. Similarly, we outperform ERNIE-Layout \cite{peng2022ernielayoutlayoutknowledgeenhanced} on Kleister NDA by 6 points.


Concerning SlideVQA \cite{tanaka2023slidevqadatasetdocumentvisual} that is based on visually rich presentations of up to 20 slides,  we obtain 2 points less than GPT-4 Vision.
On recently introduced MMLongBench-Doc \cite{ma2024mmlongbenchdocbenchmarkinglongcontextdocument} that evaluates zero-shot performance on documents as long as 400 pages, we outperform vastly larger LLMs: Mixtral 8x7B by 8 points, QWen-Plus by 6 points, and  LVLMs: Claude-3 Opus by 9 points, InternVL by 11 points. Better performance was attained by models such as Gemini 1.5 Pro and GPT-4 Vision Omnia, which are believed to have hundreds of times more parameters. Whereas it was the only task considered in zero-shot setup, please note Section~\ref{sec:finetuning} studies how the performance of our model improves compared to GPT-4o given several annotated documents.

Finally, given that three of datasets considered under this category contain labeled positions of answers within documents, we can investigate how the model's performance changes depending on the evidence location. The results shown in Figure~\ref{fig:context} indicate the \textit{primacy bias}, with the highest scores achieved when relevant information appears at the beginning of the input \cite{lostinthemiddle}.




\paragraph{Single-page.} In benchmarks involving single-page excerpts from multi-page documents or standalone images with limited input length, our model shows promising results. While Arcitc-TILT improved by 2 points over TILT on DocVQA \cite{mathew2021docvqadatasetvqadocument} and outperformed GPT-4V, it particularly excels in the newly introduced VQA-CD dataset, which includes invoices and purchase orders, establishing state-of-the-art results \cite{souleimanmahamoud:hal-04089332}.

Although there is still a gap compared to 102B InternVL 2.0 Pro’s performance \cite{chen2024fargpt4vclosinggap}, especially in non-business InfographicsVQA \cite{mathew2021infographicvqa}, our achievements highlight significant advancements in handling multi-modal inputs.

\subsection{Layout-Aware Summarization}
To supplement VQA and KIE results, we examine how Arctic-TILT exploits layout information and captures long-range dependencies in the LoRaLay collection of summarization tasks where, in contrast to the majority of similar datasets, input is not plain text but a scientific document with rich structure \cite{nguyen2023loralaymultilingualmultimodaldataset}.

Results presented in Table~\ref{tab:results} show that even though, in contrast to the previous SOTA, we had no pretraining objective explicitly designed for the summarization task, we could outperform the best model by a few points on both  ArXiv-Lay and PubMed-Lay.

\begin{figure}[t]
    \centering
    \includegraphics[width=\linewidth]{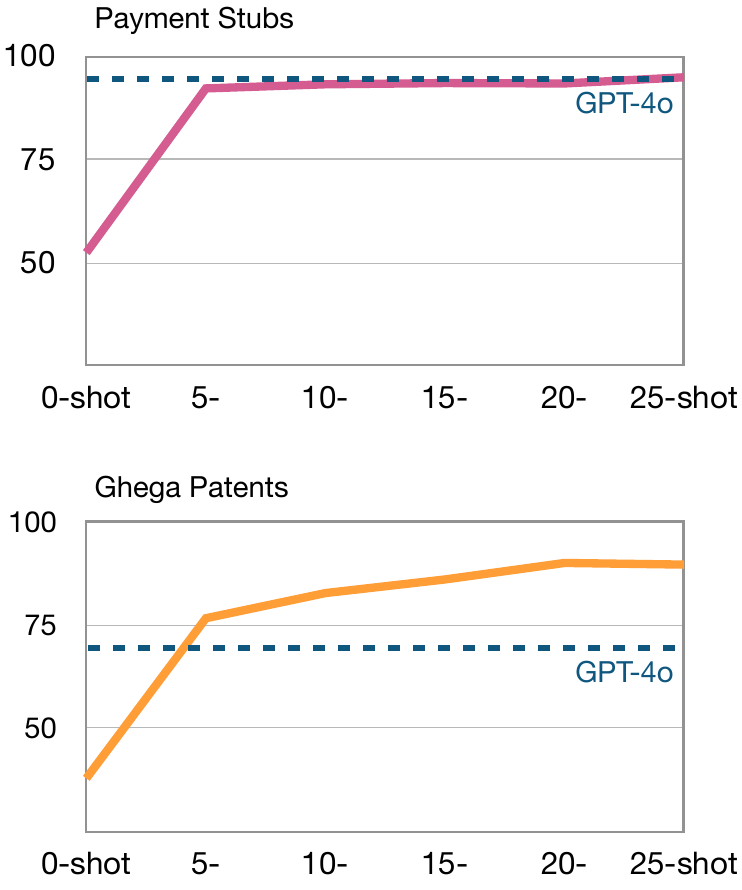}
    \caption{Improvement of Arctic-TILT zero-shot accuracy given fine-tuning on up to 25 annotated documents. Zero-shot performance of GPT-4o for comparison.}
    \label{fig:finetuning}
\end{figure}

\subsection{Adapting to Novel Use Cases}\label{sec:finetuning} 

Some optimizations introduced in Arctic-TILT aim to improve training performance under the minimal memory regime. These capabilities enable further improvement of the model in a production environment, especially when encountering out-of-domain examples or novel use cases, and appear vital in line with previous works, which have shown that smaller LLMs can outperform larger, prompted models assuming we allow fine-tuning \cite{zhao2024loraland310finetuned,bucher2024finetunedsmallllmsstill}.

We study how the zero-shot accuracy of Arctic-TILT increases, given fine-tuning on up to 25 annotated documents from holdout datasets. In particular, we rely on Ghega patents \cite{medvet2011probabilistic} and a private dataset of payment stubs and compare the model's performance to GPT-4o (refer to Appendix~\ref{appendix:private_datasets} for dataset's details).


Results shown in Figure~\ref{fig:finetuning} demonstrate that Arctic-TILT quickly approaches the accuracy of GPT-4o with as few as five annotated examples and outperform it given slightly more. These findings support the argument for employing specialized, `smaller' LLMs over a single general-purpose model in production, emphasizing the solution's cost-effectiveness and adaptability.


\subsection{Confidence Calibration}

\begin{figure}[t]
    \includegraphics[width=0.92\linewidth]{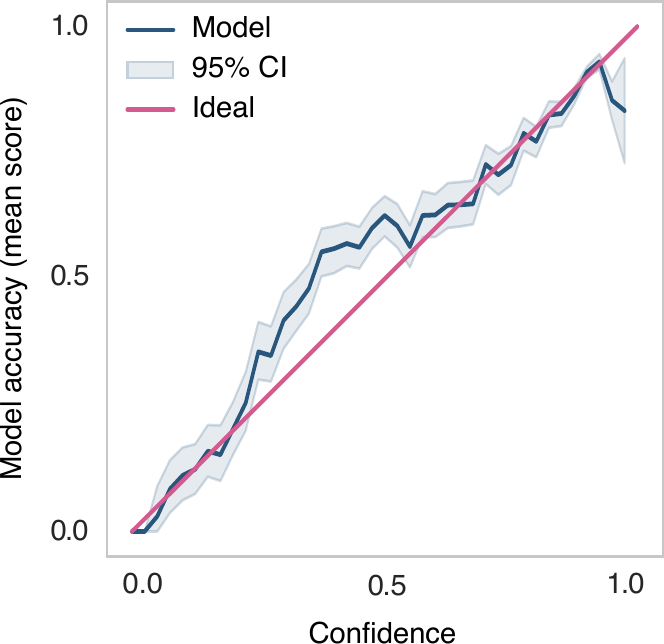}
    \caption{Arctic-TILT calibration.}
    \label{fig:calibration}
\end{figure}

Following the postulate of \Citet{vanlandeghem2023documentunderstandingdatasetevaluation}, we evaluate the \textit{Expected Calibration Error} (ECE) and \textit{Area Under the Risk-Coverage Curve} (AURC) on the DUDE dataset. Each answer's confidence is computed from a list of per-token scores. In contrast to some previous works, we take the minimum score in the list rather than the geometric mean as we found it empirically superior.

Obtained results show exceptional calibration and confidence assessment, achieving a state-of-the-art ECE of 7.6, significantly improving upon the previous best of 19.0. This suggests a closer alignment between model confidence and accuracy. Additionally, our AURC of 25.3, which surpasses the previous best of 44.0, demonstrates that our model can effectively discriminate between correct and incorrect predictions. It also shows our model's ability to appropriately assign low-confidence to predictions demanding additional human review.

To explore the landscape beyond a single dataset, we provide results on 18k data points sampled from fourteen private and public datasets in Figure~\ref{fig:calibration}. The analysis confirms low ECE and indicates that the Arctic-TILT confidence score is well calibrated as the accuracy (mean score) follows the diagonal $y=x$ on the calibration plot.






\subsection{Computational Efficiency}

\begin{figure}[t]
    \centering
    \includegraphics[width=\linewidth]{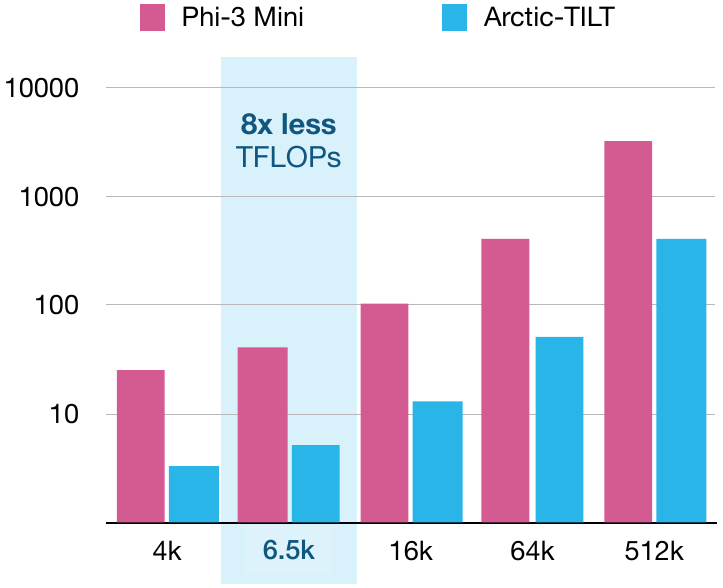}
    \caption{Arctic-TILT's computational efficiency (TFLOPs, lower is better) compared to Phi-3 Mini on VQA/KIE given inputs ranging from 4k to 512k tokens. 
    }
    \label{fig:flops}
\end{figure}

The imperative for businesses to rapidly and efficiently process substantial document volumes calls for models that maximize throughput while also maximizing operational efficiency.

To address this aspect of the model, we analyze the inference floating point operations per second (TFLOP) required for Arctic-TILT compared to Phi-3 Mini \cite{abdin2024phi3technicalreporthighly}, an example of a decoder-only model featuring 3.8B parameters and optimized by resorting to the attention sliding window. The latter was selected as a well-known reference model concerning the limited memory and compute regime we aim at, though it is not capable of achieving satisfactory accuracy on Document Understanding tasks.

Results presented in Figure~\ref{fig:flops} indicate that Arctic-TILT consistently demands lower TFLOP across all context lengths for our primary use case of VQA/KIE,\footnote{We assume the output of 8 tokens, which is longer than the average target length of evaluation datasets mentioned in Section~\ref{sec:qa_kie}.} reflecting its smaller parameter size. Importantly, concerning the input of 6.5k tokens, the mean input length for VQA/KIE tasks considered before, we require 8$\times$ less operations.

\section{Summary} 

We have introduced the Arctic-TILT model, which addresses TILT's limitations in handling multi-modal input, suboptimal training procedure, and maximum context length. By analyzing the results and considering the cost-efficiency of the designed solution, we provided practical insights into designing capable, lightweight models for the industry. In particular, we:

\begin{itemize}[leftmargin=*]
    \item established state-of-the-art performance on seven benchmarks demanding text, vision, and layout comprehension;
    \item demonstrated that within the industrial applications setting and while keeping the parameter count below 1B, one could achieve performance better or comparable to vastly larger LLMs and LVLMs;
    \item presented a novel modality fusion mechanism inspired by tensor product representations, and have shown how effectively apply it across the transformer encoder;
    \item demonstrated how, with well-designed attention sparsity patterns and numerous other optimizations, consume extensive input sequences during training and inference, given a single cost-efficient GPU, while maintaining competitive accuracy of the model;
   \item provided insights that can be applied to design future generations of multimodal models, particularly for visually-rich document processing.

\end{itemize}

\noindent Our work illustrates that strategic design and optimization can rival the capabilities of larger, more resource-intensive models.

\section*{Acknowledgements}
We express our sincere gratitude to Tomasz Dwojak and Daniel Campos for their feedback on the manuscript and suggestions that have greatly enhanced the quality of this work.
We also extend our thanks to Łukasz Słabinski, Michał Gdak, Tomasz Stanisławek, Nikolai Scholz, and Vivek Raghunathan, whose support and guidance as managers have been helpful throughout this research.
Finally, we thank Rafał Kobiela for his assistance with the cloud infrastructure.


\bibliography{custom}
\bibliographystyle{acl_natbib}

\DeclareRobustCommand{\VAN}[3]{#3}

\onecolumn
\clearpage
\appendix

\begin{lstlisting}[language=Python,label=lst:fusion,caption=Complete Arctic-TILT modality fusion module.]
class TiltLayerNorm(nn.Module):
    """
    This is essentially the T5 modification of layer norm, referred to as RMS norm.

    Args:
        dim: the dimension of vectors to be normalized, i.e. the last dimension of the input tensor
        eps: small positive value added to computed second moment for numerical stability
    """

    def __init__(self, dim: int, eps: float = 1e-6) -> None:
        super().__init__()
        self.w = nn.Parameter(torch.ones(dim))
        self.eps = eps
        self.init_weights()

    def forward(self, inp: Tensor) -> Tensor:
        dtype = inp.dtype
        x = inp.to(torch.float32)
        squared_norm = x.pow(2).mean(dim=-1, keepdim=True)
        x = x * torch.rsqrt(squared_norm + self.eps)
        return self.w * x.to(dtype)

    def init_weights(self, factor: float = 1.0) -> None:
        self.w.data.fill_(factor * 1.0)


class TiltPostFusionModule(nn.Module):
    """
    Introduced in the Arctic-TILT paper.

    Args:
        d_model: dimension of input vectors
        dropout: probability of dropout applied to input embeddings
        layer_norm: the module responsible for input embeddings
    """

    def __init__(self, d_model: int, dropout: float, layer_norm: TiltLayerNorm):
        super().__init__()
        self.layer_norm = layer_norm
        self.to_v = nn.Linear(d_model, d_model, bias=False)
        self.to_out = nn.Linear(d_model, d_model, bias=False)
        self.to_r = nn.Linear(d_model, d_model, bias=False)
        self.dropout = nn.Dropout(dropout)

    def forward(self, text_queries: Tensor, image_queries: Tensor) -> Tensor:
        """
        Compute module's forward pass.

        Args:
            text_queries (Tensor): Tensor representing the primary input in the fusion, which is text-based, or mixed.
            image_queries (Tensor):  Tensor representing the secondary input in the fusion, which is image-based.
        """
        bs, l, d = text_queries.shape
        inputs = torch.stack([text_queries, image_queries], dim=-2)
        inputs = inputs.view(bs * l, 2, d)
        normed_inputs = self.dropout(self.layer_norm(inputs))
        normed_primary_input = normed_inputs[:, 0]
        out: Tensor = self.to_v(normed_inputs.sum(-2))
        out = out + out * self.to_r(normed_primary_input)
        out = self.to_out(out)
        out = out.view(bs, l, d)
        return text_queries + out
\end{lstlisting}

\section{Why TILT as a Starting Point?}
We argue that the effectiveness of the DU model depends primarily on its ability to understand specific document formats and structures in the most document-native way possible, which can only be guaranteed by equipping the model with layout-aware architectural biases as early as possible. 

Though a number of large vision-only models have been proposed \cite{kim2022donut,davis2022endtoenddocumentrecognitionunderstanding,Lee2022Pix2StructSP}, smaller models with an explicit OCR step still outperform them. Notably, even GPT-4 Vision benefits from the availability of OCR-recognized text \cite{borchmann2024notesapplicabilitygpt4document}.
Although document intelligence requires visual features (e.g., to recognize checkboxes, signatures, text colors, and formatting), the text and its spatial arrangement are most important. This necessitates models with heavy textual and lightweight visual encoders, such as TILT.

Secondly, the imperative for businesses to rapidly and efficiently process substantial document volumes calls for models that maximize throughput while also maximizing operational efficiency. Smaller, specialized models, tailored for such tasks, often surpass their larger LLM counterparts, which struggle to meet these criteria due to their higher computational demands and processing times. The motivation for these is not only practical, as regulations such as GDPR, CCPA, or Chinese digital laws may require specific types of information to be processed locally. This need is fulfilled with smaller, specialized models that can be deployed on broadly available GPUs and thus are not restricted to a handful of regions.

The original TILT offers impressive performance despite keeping the number of parameters below 1B because of a well-balanced parameter budget and relying on encoder-decoder architecture, which, despite lower popularity compared to decoder-only models, offers better quality in compute-matched setups \cite{2020t5,chowdhery2022palmscalinglanguagemodeling,wang2022languagemodelarchitecturepretraining,tay2023ul2unifyinglanguagelearning}.
Besides, we prefer them because achieving optimal attention sparsity patterns is more straightforward with separate encoder and decoder modules.

The encoder-decoder model with a sizeable textual backbone and small visual encoder, equipped with layout architectural bias that has previously established state-of-the-art results, appears a viable starting point for building a modern DU system.

\section{Datasets for Supervised Finetuning}\label{appendix:dataset}
Training of Arctic-TILT included SFT phase on twelve publicly available and five in-house annotated datasets. The first group included Kleister Charity, Kleister NDA \cite{kleister-10.1007/978-3-030-86549-8_36}, CHART-Infographics \cite{CHART_Infographics_9956289}, 
DeepForm$^\bigstar$ \cite{borchmann2021due}, DocVQA \cite{mathew2021docvqadatasetvqadocument}, DUDE \Citep{vanlandeghem2023documentunderstandingdatasetevaluation}, FUNSD \cite{funsd_8892998}, InfographicVQA \cite{mathew2021infographicvqa}, SQuAD 2.0 \cite{sqaud2_rajpurkar2018knowdontknowunanswerable}, TAT-DQA \cite{tat_dqa_zhu2022towards}, VQA-CD \cite{vqa_cd-10.1007/978-3-031-06555-2_44}, and VQAonBD \cite{vqa_bd-10.1007/978-3-031-41679-8_26}.

\begin{figure}[t]
    \centering
    \includegraphics[width=\linewidth]{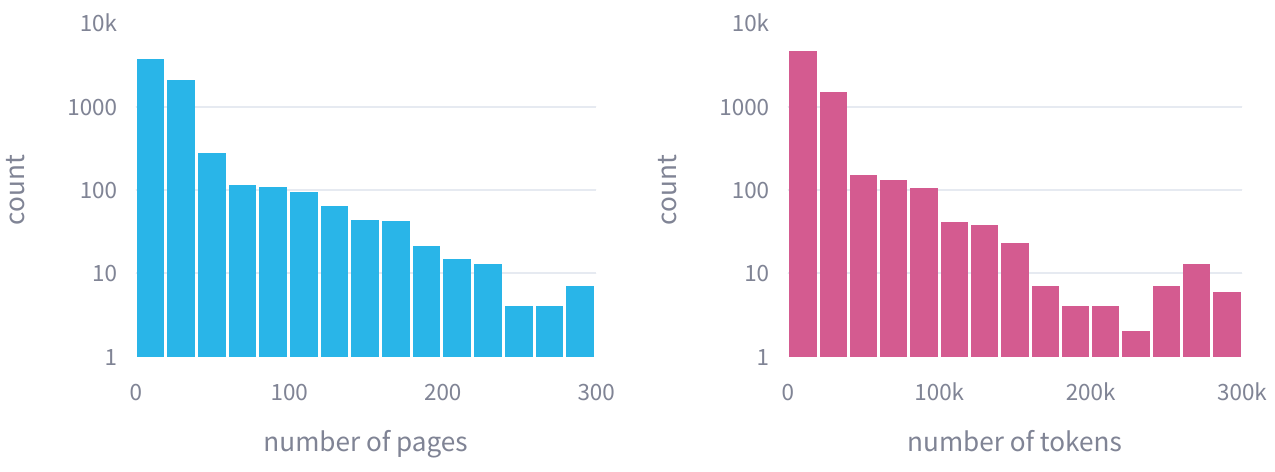}
    \caption{Lengths of documents included in five private datasets (number of tokens and pages).}
    \label{fig:private}
\end{figure}

Private datasets were based on QA annotations of IRS990 forms, insurance reports, company annual reports, synthetic invoices, and charity annual reports. To give the research community a grasp on the characteristics of this collection, we provide the most important statistics and examples of questions in Figure~\ref{fig:private} and Table~\ref{tab:private}.

\begin{table*}[h]
   \small
   \caption{Outline of private datasets used for SFT.} 
   \label{tab:private}
   \centering
   \bgroup
   \def\arraystretch{1.5}
   \begin{tabular}{l p{0.5\linewidth} r r}
   \toprule
   Dataset & Sample Questions & Documents & Annotations \\ 
   \midrule
   IRS990 & What is the percentage of public support in the year of the report? What is the sum of the total liabilities in US dollars? What is the Employer Identification Number? & 3,097 & 38,025 \\
   Insurance Reports & What was the value of total premiums written in the Surplus \& Self-Procured category in 2016? Who is the director of the Alaska Division of Insurance? For whose contributions were tax credits claimed in 2015? & 50 & 1,702 \\
   Company Annual Reports & What is the name of the chief executive officer? What is the total net income for report year? What is the tier 1 capital ratio? & 648 & 3,522 \\
   Synthetic Invoices & What is the number of items on the invoice? What is the total net amount of the item described on the invoice? What is the description of the item of the transaction? & 2,707 & 17,274 \\
   Charity Annual Reports & What is the independent auditor's name? What are the charity's total funds in the bank and in hand? What is the name of the organisation's chairman?  & 161	& 4,025 \\
   \bottomrule
   \end{tabular}
   \egroup
\end{table*}

\section{Used Hyperparameters and their Optimization}\label{sec:hparams}
\paragraph{Chunking setup.} We studied the size of the attention block, as well as the overlap size of consecutive blocks. To our surprise, the best setup for inference was 1024 tokens attention size with no overlap, and these conclusions are independent of the setup overlap/attention size during training. 

\begin{figure}[h]
    \centering
    \includegraphics[width=0.9\linewidth]{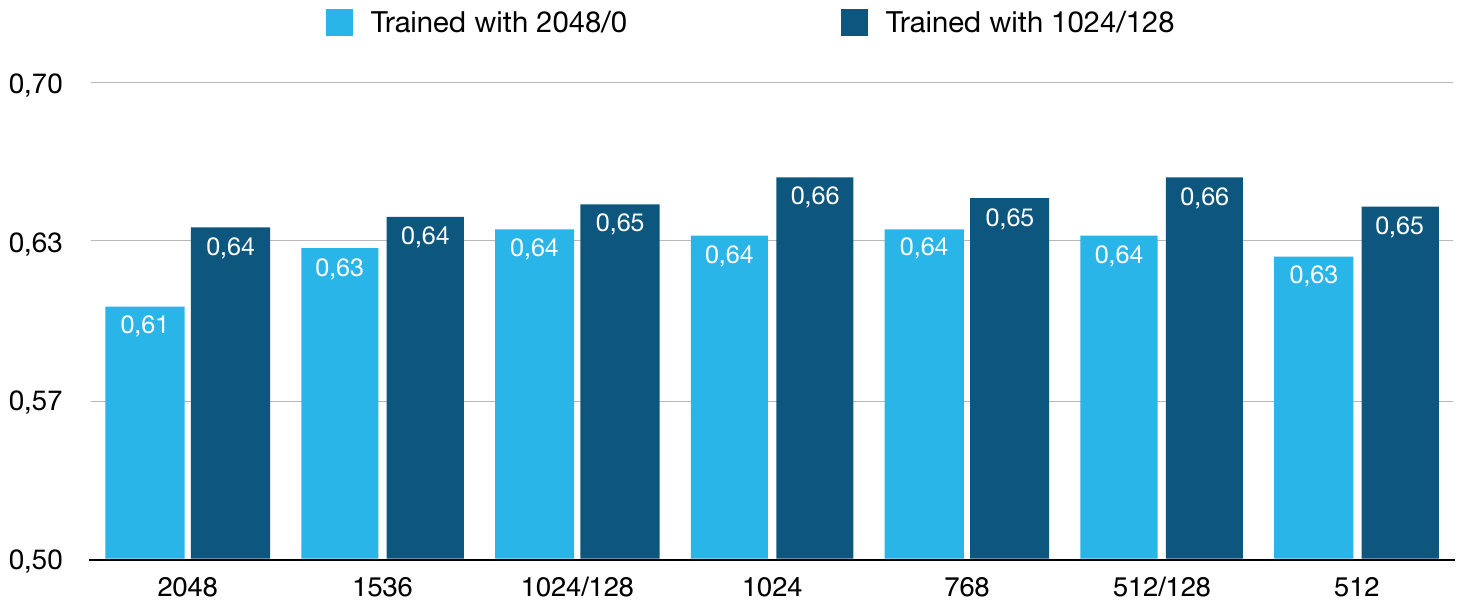}
    \caption{Impact of chunk size and overlap size (chunk/overlap) on a downstream inference for two models, assuming in-house dataset of business use cases. We observe no positive impact of overlap for sufficiently long input sequences, such as 1024 tokens.}
    \label{fig:chunk-size-and-overlap}
\end{figure}

\paragraph{Learning rate scheduling and precision.} We observed a non-trivial inference between the two hyper-parameters. Compared to \textit{fp32} pretraining, \textit{bf16} pretraining with more aggresive learning rate scheduling was able to catch up, and with same learning rate scheduling was observably worse. We ended up with using \textit{cosine\_luh} scheduler with 1\% training steps with constant learning rate of 1e-3 (warm-up), followed by 89\% training steps with linear decay down to 2e-4, followed by cosine scheduling for the remaining 10\% steps decaying to 5e-5. Same observations were drawn during fine-tuning. 


\paragraph{Training protocol.}
The finetuning phase's hyperparameters are set as 100k steps at batch size 128 with the \textit{AdamWScale} optimizer. We set loss reduction to mean and weight decay to $1\mathrm{e}{-5}$. Additionally, we used case augmentation of the whole triple consisting of the document, question, and answer. Specifically, if we detect that the document is not already cast to upper or lowercase, we create an augmented version of the three-tuple question-document-answer by casting them all to that case, similarly to \citet{tilt}. This means that there are up to three versions of each data point, such as the original one, uppercase, and lowercase.

\paragraph{Downstream tasks evaluation.} 
For downstream task evaluation on benchmarks providing trainset (DocVQA, MP-DocVQA, DUDE, Kleister Charity, Kleister NDA, SlideVQA, InfographicsVQA, VQA-CD) we performed additional training with Optuna \cite{optuna_2019} hyperparameter tuning. We performed 10-40 studies optimizing the following hyperparameters:

\begin{itemize}
    \item \textit{case augmentation} (on, off) -- augment dataset with lowercased/uppercased version of training samples, in case they are statistically distinguishable;
    \item \textit{answer variants sampling} (on, off) -- for questions with multiple versions of the correct answer (e.g. \texttt{100}, \texttt{\$100}), we either pick the same, or sample variant per epoch;
    \item \textit{dropout} sampled with uniform distribution from the interval \((0, 0.2)\);
    \item \textit{weight decay} sampled with log-uniform distribution from the interval \((1\mathrm{e}{-6}, 1\mathrm{e}{-2})\);
    \item \textit{learning rate} sampled with log-uniform distribution from the interval \((1\mathrm{e}{-4}, 5\mathrm{e}{-3})\).
\end{itemize}

\section{Finetuning Study}\label{appendix:private_datasets}

\paragraph{GPT-4o baseline.} Following the findings of \citet{borchmann2024notesapplicabilitygpt4document}, we assume input images of 2048px along longer dimensions (usually height) and similar prompts. The latter were subject to further per-dataset optimization to cover the convention used in considered datasets (final form presented in Table~\ref{tab:prompts}).

\paragraph{Payment Stubs.} The private dataset used for evaluation consists of American payment stubs, i.e., documents obtained by an employee regarding the salary received. The test split contains 39 documents with 448 annotations. Since all come from different companies, their layouts differ significantly. Questions aim to extract employee and employer names, dates, addresses and information from payment tables, where each row consists of payment type, hours worked, and payment amount, e.g., `What is the name of the US state of the employee's address?' or `When does the pay period finish?'

\begin{table*}[t]
   \small
   \caption{Final prompts used for GPT-4o baselines.} 
   \label{tab:prompts}
   \centering
   \bgroup
   \def\arraystretch{1.5}
   \begin{tabular}{l p{0.8\linewidth}}
   \toprule
   Dataset & Prompt \\ 
   \midrule
   Payment Stubs & Replace \texttt{[ANSWER]} with a value in the template given question and document. $\hookleftarrow$ Question: \texttt{[TEXT]} $\hookleftarrow$  Template: Based on the context, the answer to the question would be "\texttt{[ANSWER]}". $\hookleftarrow$ $\hookleftarrow$ Normalize amounts to two decimal places, without thousand separator and without dollar sign. $\hookleftarrow$ Normalize states using postal abbreviations, e.g., TX or NJ. \\
   Ghega Patents & Replace \texttt{[ANSWER]} with a value in the template given question and document. $\hookleftarrow$ Question: \texttt{[TEXT]} $\hookleftarrow$  Template: Based on the context, the answer to the question would be "\texttt{[ANSWER]}". $\hookleftarrow$ $\hookleftarrow$ Normalize dates to YYYY-MM-DD format except question about priority which should remain similar to "DD.MM.RRRR (country code) (optional number)." \\
   \bottomrule
   \end{tabular}
   \egroup
\end{table*}

\section{Contributions}\label{sec:contributions}

\paragraph{ŁB.} Performing early-stage architecture ablations, writing most of the paper and preparing figures, final fusion by TP module design and related ablation studies, implementation of GPT-4o baseline for Arctic-TILT SFT, analysis of results on public benchmarks, overseeing initial model sparsification experiments, study of long context utilization, self-supervised pretraining of the model.

\paragraph{ŁD.} Various contributions related to configurations and automatizations of experiments.

\paragraph{PD.} Optimization of various data pipelines (image processing, loading, metric computation). Datasets updates and modifications (main focus on increasing loading speed and OCR correctness).

\paragraph{ŁG.} Technical leadership and participation in the codebase implementation, performance optimization, and attention sparsity efforts. Writing parts of the paper.

\paragraph{AG.} Efficient implementation of attention sparsity. Contributions to codebase implementation and memory optimizations. Preparation and cleaning of some datasets. Experiments with model sizes and architectures.

\paragraph{PH.} Major contributions to memory- and compute-efficiency, including evaluation of long context approaches with a theoretical memory model, implementing nested checkpointing and mixed-precision, and empirically evaluating the complete solution's performance and memory usage characteristics.

\paragraph{WJ.} Leading SFT efforts, including fine-tuning the model's final version, experiments with hyperparameters, training protocols, and dataset composition. Model performance improvements. Various contributions in the model and training code. Preparation and cleaning of some datasets.

\paragraph{PJ.} Memory fragmentation handling. Implementation and creation of semi-synthetic long document Needle-in-a-Haystack benchmark used in early experiments. Data management, curating final training and performing a few downstream evaluations (DUDE, VQA\_CD). Writing parts of the paper.

\paragraph{DJ.}
Leading efforts to increase the context length from 25 to 500 pages (conceptualization, brainstorming, planning, guidance). Conducting initial memory and throughput experiments, as well as final stress tests and quality experiments for very large context lengths. Implementation of CPU offloading. Performing few-shot finetuning experiments. Delivering results for SlideVQA, Kleister Charity, and Kleister NDA datasets. 

\paragraph{PL.} Implemented TP fusion and conducted ablation studies focusing on module placement. Devised enhanced training protocol, performed hyperparameter tuning, and contributed to various model improvements. Performed experimental evaluation on DocVQA and InfographicsVQA.

\paragraph{GN.} Preparation and management of datasets, the idea behind creating semi-synthetic long documents, automation of data processing pipeline, conducting experiments and analyzing the results with long documents.

\paragraph{JO.} Selection and improvements of the training datasets (analysis of data quality, filtering the data, fixing quality issues), optimizations of image encoder and data processing.

\paragraph{MP.} Performing early-stage architecture ablations (researching, implementing, and studying effects) that lead to co-authoring TP fusion (i.e., proposing initial attn-based version, module placement study, and fusion in every layer). Leading efforts in writing parts of the paper (technical optimizations, training, related works, analysis, structuring and rewriting).

\paragraph{KS.} The idea behind attention sparsification, ablation studies of various approaches, implementation of required prototypes, and analysis of the results.

\paragraph{MT.} Training loop optimization (in terms of processing time and data efficiency), performing downstream evaluations (DocVQA, MP-DocVQA, MMLongBench-Doc), dataset preparation and cleaning, error analysis, and organization of work.

\paragraph{AZ.} Various contributions to the development of the codebase. 

\end{document}